\documentclass[10pt,journal,compsoc]{IEEEtran}
\usepackage{graphicx}
\usepackage{nicefrac}
\usepackage{rotating}
\usepackage{fancybox}
\usepackage{enumerate}
\usepackage{amsfonts}
\usepackage{amsmath}
\usepackage{comment}
\usepackage{amssymb}
\usepackage{amscd,bm,amsbsy}
\usepackage{setspace}

\newcommand{\by}{\boldsymbol{y}}
\newcommand{\bL}{\boldsymbol{L}}
\newcommand{\bD}{\boldsymbol{D}}

\newcommand{\bW}{\boldsymbol{W}}

\newcommand{\bM}{\boldsymbol{M}}
\newcommand{\bF}{\boldsymbol{F}}

\newcommand{\bbf}{\boldsymbol{f}}

\newcommand{\bY}{\boldsymbol{Y}}

\newcommand{\field}[1]{\mathbb{#1}}
\newcommand{\R}{\field{R}}

\newtheorem{theorem}{Fact}
\newtheorem{corollary}{Corollary}
\DeclareMathOperator{\trace}{trace}

\DeclareMathOperator{\MA}{MA}
\DeclareMathOperator{\diss}{diss}
\DeclareMathOperator{\simm}{sim}
\DeclareMathOperator{\anc}{anc}

\newcommand{\scA}{\mathcal{A}}
\newcommand{\scC}{\mathcal{C}}
\newcommand{\bscA}{\boldsymbol{\mathcal{A}}}
\newcommand{\bscC}{\boldsymbol{\mathcal{C}}}

\newcommand{\bscI}{\boldsymbol{\mathcal{I}}}
\newcommand{\bscL}{\boldsymbol{\mathcal{L}}}
\newcommand{\wY}{\widetilde{Y}}
\newcommand{\wbY}{\widetilde{\bY}}
\renewcommand{\bar}{\overline}
\newcommand{\TP}{\mathrm{TP}}
\newcommand{\TN}{\mathrm{TN}}
\newcommand{\FP}{\mathrm{FP}}
\newcommand{\FN}{\mathrm{FN}}
\newcommand{\Prec}{\mathrm{Prec}}
\newcommand{\Rec}{\mathrm{Rec}}
\newcommand{\Fmax}{F_{\max}}

\ifCLASSOPTIONcompsoc
  \usepackage[nocompress]{cite}
\else
  \usepackage{cite}
\fi
\ifCLASSINFOpdf
\else
\fi
\begin{document}
%
% paper title
% Titles are generally capitalized except for words such as a, an, and, as,
% at, but, by, for, in, nor, of, on, or, the, to and up, which are usually
% not capitalized unless they are the first or last word of the title.
% Linebreaks \\ can be used within to get better formatting as desired.
% Do not put math or special symbols in the title.
\title{
%Multitask Label Propagation with Dissimilarity Measures (da rivedere)
Multitask Protein Function Prediction \\ Through Task Dissimilarity
}
%
%
% author names and IEEE memberships
% note positions of commas and nonbreaking spaces ( ~ ) LaTeX will not break
% a structure at a ~ so this keeps an author's name from being broken across
% two lines.
% use \thanks{} to gain access to the first footnote area
% a separate \thanks must be used for each paragraph as LaTeX2e's \thanks
% was not built to handle multiple paragraphs
%
%
%\IEEEcompsocitemizethanks is a special \thanks that produces the bulleted
% lists the Computer Society journals use for "first footnote" author
% affiliations. Use \IEEEcompsocthanksitem which works much like \item
% for each affiliation group. When not in compsoc mode,
% \IEEEcompsocitemizethanks becomes like \thanks and
% \IEEEcompsocthanksitem becomes a line break with idention. This
% facilitates dual compilation, although admittedly the differences in the
% desired content of \author between the different types of papers makes a
% one-size-fits-all approach a daunting prospect. For instance, compsoc 
% journal papers have the author affiliations above the "Manuscript
% received ..."  text while in non-compsoc journals this is reversed. Sigh.

\author{Marco~Frasca,
        Nicol\`o~Cesa Bianchi% <-this % stops a space
\IEEEcompsocitemizethanks{\IEEEcompsocthanksitem M.~Frasca and N.~Cesa Bianchi are with the Dipartimento di Informatica, Universit\`a degli Studi di Milano, Via Comelico 39 Milano, 20137,
Italy.\protect\\
% note need leading \protect in front of \\ to get a newline within \thanks as
% \\ is fragile and will error, could use \hfil\break instead.
E-mail: $\{$marco.frasca, nicolo.cesa-bianchi$\}$@unimi.it
}% <-this % stops an unwanted space
\thanks{Manuscript received April 19, 2005; revised August 26, 2015.}}

% note the % following the last \IEEEmembership and also \thanks - 
% these prevent an unwanted space from occurring between the last author name
% and the end of the author line. i.e., if you had this:
% 
% \author{....lastname \thanks{...} \thanks{...} }
%                     ^------------^------------^----Do not want these spaces!
%
% a space would be appended to the last name and could cause every name on that
% line to be shifted left slightly. This is one of those "LaTeX things". For
% instance, "\textbf{A} \textbf{B}" will typeset as "A B" not "AB". To get
% "AB" then you have to do: "\textbf{A}\textbf{B}"
% \thanks is no different in this regard, so shield the last } of each \thanks
% that ends a line with a % and do not let a space in before the next \thanks.
% Spaces after \IEEEmembership other than the last one are OK (and needed) as
% you are supposed to have spaces between the names. For what it is worth,
% this is a minor point as most people would not even notice if the said evil
% space somehow managed to creep in.

% The paper headers
\markboth{Journal of \LaTeX\ Class Files,~Vol.~14, No.~8, August~2015}%
{Shell \MakeLowercase{\textit{et al.}}: Bare Demo of IEEEtran.cls for Computer Society Journals}
% The only time the second header will appear is for the odd numbered pages
% after the title page when using the twoside option.
% 
% *** Note that you probably will NOT want to include the author's ***
% *** name in the headers of peer review papers.                   ***
% You can use \ifCLASSOPTIONpeerreview for conditional compilation here if
% you desire.

% The publisher's ID mark at the bottom of the page is less important with
% Computer Society journal papers as those publications place the marks
% outside of the main text columns and, therefore, unlike regular IEEE
% journals, the available text space is not reduced by their presence.
% If you want to put a publisher's ID mark on the page you can do it like
% this:
%\IEEEpubid{0000--0000/00\$00.00~\copyright~2015 IEEE}
% or like this to get the Computer Society new two part style.
%\IEEEpubid{\makebox[\columnwidth]{\hfill 0000--0000/00/\$00.00~\copyright~2015 IEEE}%
%\hspace{\columnsep}\makebox[\columnwidth]{Published by the IEEE Computer Society\hfill}}
% Remember, if you use this you must call \IEEEpubidadjcol in the second
% column for its text to clear the IEEEpubid mark (Computer Society jorunal
% papers don't need this extra clearance.)

% use for special paper notices
%\IEEEspecialpapernotice{(Invited Paper)}

% for Computer Society papers, we must declare the abstract and index terms
% PRIOR to the title within the \IEEEtitleabstractindextext IEEEtran
% command as these need to go into the title area created by \maketitle.
% As a general rule, do not put math, special symbols or citations
% in the abstract or keywords.
\IEEEtitleabstractindextext{%
\begin{abstract}
Automated protein function prediction is a challenging problem with distinctive features, such as the hierarchical organization of protein functions and the scarcity of annotated proteins for most biological functions. We propose a multitask learning algorithm addressing both issues. Unlike standard multitask algorithms, which use task (protein functions) similarity information as a bias to speed up learning, we show that dissimilarity information enforces separation of rare class labels from frequent class labels, and for this reason is better suited for solving unbalanced protein function prediction problems. We support our claim by showing that a multitask extension of the label propagation algorithm empirically works best when the task relatedness information is represented using a dissimilarity matrix as opposed to a similarity matrix. Moreover, the experimental comparison carried out on three model organism shows that our method has a more stable performance in both ``protein-centric'' and ``function-centric'' evaluation settings.
\end{abstract}

% Note that keywords are not normally used for peerreview papers.
}

% make the title area
\maketitle

% To allow for easy dual compilation without having to reenter the
% abstract/keywords data, the \IEEEtitleabstractindextext text will
% not be used in maketitle, but will appear (i.e., to be "transported")
% here as \IEEEdisplaynontitleabstractindextext when the compsoc 
% or transmag modes are not selected <OR> if conference mode is selected 
% - because all conference papers position the abstract like regular
% papers do.
\IEEEdisplaynontitleabstractindextext
% \IEEEdisplaynontitleabstractindextext has no effect when using
% compsoc or transmag under a non-conference mode.

% For peer review papers, you can put extra information on the cover
% page as needed:
% \ifCLASSOPTIONpeerreview
% \begin{center} \bfseries EDICS Category: 3-BBND \end{center}
% \fi
%
% For peerreview papers, this IEEEtran command inserts a page break and
% creates the second title. It will be ignored for other modes.
\IEEEpeerreviewmaketitle

\IEEEraisesectionheading{\section{Introduction}\label{sec:introduction}}
\IEEEPARstart{T}{he} constant increase in the volume and variety of publicly available genomic and proteomic data is a characteristic trait of modern biomedical sciences. A fundamental problem in this area is the assignment of functions to biological macromolecules, especially proteins. Indeed, the accurate annotation of protein function would also have great biomedical and pharmaceutical implications, since several human diseases have genetic causes. While molecular experiments provide the most reliable annotation of proteins, their relatively low throughput and restricted scope have led to an increasing role for automated function prediction (AFP).
AFP is characterized by unbalanced functional classes with rare positive instances. Moreover, since only positive membership to functional classes is usually assessed, negative instances are not uniquely defined, and different approaches to choose them have been proposed~\cite{Youngs13,Frasca15Wirn,Mostafavi09}. Other peculiarities of AFP include: (1) the need of integrating several heterogeneous sources of genomic, proteomic, and transcriptomic data in order to achieve more accurate predictions~\cite{Mostafavi10,Frasca15c}; (2) the presence of multiple labels and dependencies among class labels; (3) the hierarchical structure of functional classes (a direct acyclic graph for the \textit{Gene Ontology} GO~\cite{GO00}, a forest of trees for the \textit{FunCat} taxonomy~\cite{Ruepp04}) with different levels of specificity.

Recently, two international challenges for Critical Assessment of Functional Annotation, (CAFA~\cite{Radivojac13} and CAFA2~\cite{CAFA2}) were organized to evaluate computational methods that automatically assign protein functions. In particular, CAFA2 emphasized the need for multilabel or structured-output learning algorithms for predicting a set of terms, or a subgraph of the GO ontology for a given protein. In this work we mainly focus on this problem, whose solution however requires paying attention also to the other aspects of AFP.

%Usually methods for AFP problem are two-class classifiers, which predict for each class whether genes are members or not of the functional class, or rankers, that is they provide for each gene a real score such that the higher the score the stronger the clue of gene membership to the class.
Several approaches to the predicton of protein functions were proposed in the literature, including sequence-based~\cite{Martin04,Hawkins09,Juncker09} and network-based methods~\cite{Vazquez2003,Sharan07,Frasca15}, structured output algorithms based on kernels~\cite{Sokolov10,Sokolov13,Mostafavi09} and hierarchical ensemble methods~\cite{Obozinski08,Guan08,Vale14c}.
In particular, the availability of large-scale networks, in which nodes are genes/proteins and edges their functional pairwise relationships, has promoted the development of several machine learning methods where novel annotations are inferred by exploiting the topology of the resulting biomolecular network.
Initially, network-based approaches relied on the so called \textit{guilt-by-association} (GBA) rule, which makes predictions assuming that interacting proteins are likely to share similar functions~\cite{Marcotte99,Oliver00,Schwikowski00}. 
Indirect neighbours were also exploited to modify the notion of pairwise-similarities among nodes by accounting for pairs of nodes connected through intermediate ones~\cite{Li10, Bogdanov11}. 
Protein functions can be also propagated through the network with an iterative process until convergence~\cite{Zhu03, Zhou04}, by tuning the amount of propagation allowed in the graph through Markov random walks~\cite{Szummer01, Azran07}, by evaluating the functional flow through the nodes~\cite{Nabieva05}, by exploiting kernelized score functions~\cite{Valentini16}, and by modelling protein memberships through Markov Random Fields~\cite{Deng04} and Gaussian Random Fields~\cite{Tsuda05,Mostafavi08}. Furthermore, methods based on the convergence of classical~\cite{Karaoz04,Bertoni11} and multi-category Hopfield networks~\cite{Frasca15homcat} were recently proposed to specifically tackle the class imbalance.

Although protein functions are clearly dependent (see, e.g., the GO functions, where parent terms include all the proteins of their children), most AFP methods described above predict biological functions independently from each other. Multitask methods, on the other hand, take advantage of existing dependencies by transferring information between related tasks, which typically leads to learning faster than algorithms trained independently on each task.

In this paper we investigate an alternative approach to multitask learning based on exploiting task dissimilarities rather than similarities. In particular, we consider two multitask extensions of a known label propagation algorithm~\cite{Zhu03}: the first extension follows a standard multitask approach based on task similarities; the second extension learns instead from task dissimilarities. Both approaches can be naturally applied to the multilabel prediction of proteins. 
The prediction tasks we consider are the GO protein functions of \textit{fly}, \textit{human}, and \textit{bacteria} model organisms. We compute different measures of similarity/dissimilarity between GO terms, taking into account both GO structure and protein annotations.
We show that the approach learning from task dissimilarities greatly helps in unbalanced tasks (by helping instances labeled with the rare class labels to be correctly classified), and does not hurt in the more balanced cases. This is a crucial point in protein function prediction, since terms better describing protein functions ---i.e., the most specific ones--- are the most unbalanced (proteins annotated with these terms are very rare). On the other hand, learning from similar tasks tends to be more effective on balanced settings.
Note that the proposed multitask extensions of label propagation do not increase the overall running time of the algorithm, allowing its application on large-sized datasets. Finally, we compare our methods with the state-of-the-art methodologies for AFP by considering both ``term-centric'' and ``protein-centric'' evaluation settings.

The paper is organized as follows. In Section~\ref{sec:problem} we formally introduce the problem and in Section~\ref{sec:methods} we describe the proposed multitask label propagation methodology. Finally, Section~\ref{sec:exp} is dedicated to the experimental validation of the method.

%%%%%%%%%%%%%%%%%%%%%%%%%%%%%%%%%%%%%%%%%%%%%%%%%%%%%%%%%%%%%%%%%%%%%%%%%%%%%%

\section{Automated Protein Function Prediction}\label{sec:problem}
The Automated protein Function Prediction (AFP) problem
can be formalized as semi-supervised learning problem on a weighted and undirected graph $G = (V, E, \bW)$, where $V=\{1,\dots,n\}$ is the set of vertices, $E \subset V\times V$ is the set of edges, and 
$\bW = \big[w_{{i}{j}}\big]_{n\times n}$ is the symmetric weight matrix, where $w_{{i}{j}}$
%represents a measure of similarity
is the weight on the edge between vertices $i$ and $j$ (we assume $w_{ii} = 0$ and $w_{ij} = 0$ for all $(i,j) \notin E$).

A set of $m$ binary classification tasks on $G$ is defined by $m$ labelings $\by^{(1)},\dots,\by^{(m)} \in \{-1,1 \}^{n}$ of the nodes in $V$, where $y^{(k)}_i$ is the label of node $i$ for task $k$. For any subset $T \subseteq \{1,\dots,n\}$ and any vector $\by = (y_1,\dots,y_n)$, we use $\by_T$ to denote the vector obtained from $\by$ by retaining only the coordinates in $T$. 
%NCB This defines the subsets $N^{(k)} = \{v_i \in V \,:\, y^{(k)}_i = -1\}$ and $P^{(k)} = \{v_i \in V \,:\, y^{(k)}_i = 1\}$ of negative and positive vertices for task $k$, respectively.
%Furthermore, a weighted undirected graph over tasks $T=(\{1, \ldots, m\}, \mathcal{E}, \bC)$ is given, where $\bC$ is a $m \times m$ symmetric matrix, which encodes the prior beliefs about the pairwise task relatedness. That is, $C_{kr} \in [0,1]$ is a measure of relatedness between tasks $k$ and $r$, and $C_{kr} = 0$ when $(k,r)\notin \mathcal{E}$.

The multitask prediction problem on the graph $G$ is then defined as follows. Given a set $S \subset V$ of training vertices and the complement set $U \equiv V \setminus S$ of test vertices, the learner must predict the test labels $\by_U^{(1)},\dots,\by_U^{(m)}$ for each task given the training labels $\by_S^{(1)},\dots,\by_S^{(m)}$ for the same tasks.

% Versione binaria
%NCB The problem consists in learning for each task $k$ a labeling $\bbf^{(k)} \in \{-1,1 \}^{n}$, which aims at correctly classifying training vertices (i.e.  $y^{(k)}_i = f^{(k)}_i$ when $v_i \in S$), and at estimating the labeling $\by^{(k)}$ on nodes in $U$, so as to consider vertices $\hat U_P^{(k)} = \{v_i \in U : f^{(k)}_i = 1\}$ as candidates for the class $U_P^{(k)} = P^{(k)}\cap U$, and the remaining vertices $\hat U_N^{(k)} = \{v_i \in U :  f^{(k)}_i = -1\}$ as candidates for the class $U_N^{(k)} = N^{(k)}\cap U$. Nevertheless, a frequent variant of this problem allows the label $f^{(k)}_i$ predicted for nodes $v_i \in U$ to be just a real value, such that the higher $f^{(k)}_i$ the more likely instance $v_i$ belongs to the class $U_P^{(k)}$. Setting an appropriate threshold allows than to obtain again binary labels. 
% Computer Society journal (but not conference!) papers do something unusual
% with the very first section heading (almost always called "Introduction").
% They place it ABOVE the main text! IEEEtran.cls does not automatically do
% this for you, but you can achieve this effect with the provided
% \IEEEraisesectionheading{} command. Note the need to keep any \label that
% is to refer to the section immediately after \section in the above as
% \IEEEraisesectionheading puts \section within a raised box.

\section{Methods}\label{sec:methods}
We first describe the standard label propagation algorithm~\cite{Zhu03,Belkin02,Kveton10} for single-task classification on graphs. This will be later extended to the multitask setting.
\subsection{Label Propagation (LP)}\label{subsec:LP}
In the single-task setting, a standard notion of \textit{regularity} of a labeling $\bbf\in\{-1,1\}^n$ on a graph $G$ is the \textit{weighted cutsize} induced by $\bbf$ and defined as follows:
\begin{equation}\label{eq:w_cutsize}
	\Gamma_G^W (\bbf) = \sum_{\substack{(i,j)\in E\\f_i\neq f_j}} w_{ij}~.
\end{equation}
The weighted cutsize can be also expressed as a quadratic form
\[
	\Gamma_G^W(\bbf) = \frac{1}{4} \bbf^{\top} L \bbf = \frac{1}{4} \sum_{(i,j) \in E} w_{ij}(f_i - f_j)^2~.
\]
The matrix $\bL = \bD -\bW$ is the \textit{Laplacian} of $G$, where $\bD$ is the diagonal matrix with entries $D_{ii} = d_{i} = \sum_j w_{ij}$. The Label Propagation algorithm minimizes the above quadratic form over real-valued (rather than binary) labels. More precisely, LP finds the unique solution of
\begin{equation}
\label{eq:objective}
	\begin{array}{l}
	{\displaystyle \min_{\bbf\in\R^n} \bbf^{\top} L \bbf}
\\
	\text{s.t.} \quad f_i = y_i  \quad i \in S~.
\end{array}
\end{equation}
The solution $\bbf_U^*$ of~(\ref{eq:objective}) is smooth on $G$. Namely, if two vertices $i,j \in U$ are connected with a large weight $w_{ij}$, then $f^*_i$ is close to $f^*_j$. Indeed, the components $i \in U$ of $f^*$ satisfy the harmonic property~\cite{Zhu03}
\[
	f^*_i=\frac{1}{d_i}\sum_j w_{ij}f^*_j~.
\] 
The vector $\bbf_U^*$ can be also written in closed form as
\begin{equation}
\label{eq:solution}
	\bbf^*_U = (\bD_{UU} -\bW_{UU})^{-1}\bW_{US}\,\bbf^*_S	
\end{equation}
where
\begin{displaymath}
 \bW = \left( \begin{array}{rl}
\bW_{UU} & \bW_{US} \\
\bW_{US}^{T} & \bW_{SS} \\
 \end{array}
\right)
\end{displaymath}
is the weight matrix partitioned in blocks to emphasize the labeled and unlabeled part of the graph (similarly for the matrix $\bD$). As the components of $\bbf^*_U$ given by~(\ref{eq:solution}) are not in $\{-1,1\}$, the final labeling produced by LP is obtained by thresholding each component $f_i^*$ for $i \in U$.

\subsection{Multitask label propagation (MTLP)}\label{sec:MTLP}
It is fairly easy to use similarity or dissimilarity information between tasks in order to generalize LP to multitask learning, while preserving the regularity of every task in the sense of~(\ref{eq:w_cutsize}).

We start by considering multitask LP based on similarity information. Suppose a $m \times m$ symmetric matrix $\bscC$ is given, where each entry $\scC_{kr} \in [0,1]$ quantifies the relatedness between tasks $k$ and $r$. Let $\bscA = \gamma \bscI_m + \bscL$ be the matrix where $\gamma > 0$, $\bscI_m$ is the $m\times m$ identity matrix, and $\bscL$ is the Laplacian of $\bscC$. The matrix $\bscA$ is symmetric and  positive definite, since $\bscA$ is diagonally dominant with positive diagonals, and thus invertible. Denote by $\bY$ the $n\times m$ label matrix whose $k$-th column is the vector $\by^{(k)}$, and by $\bF$ the $n\times m$ matrix whose $k$-th column is the vector $\bbf^{(k)}$.

\iffalse
Learning the $m$ tasks independently, means minimizing the following criterion:
\begin{equation}\label{eq:ST_labprop}
	\begin{split}
%	&\min_{\bbf\in \{-1,1 \}^{n}}\|\bbf\|_L^2\\
&\min_{\bF}\ \ \ \ \trace (\bF^T\bL\bF)\\
&\mbox{s.t.\ \ \ \ \ } F_{ik} = Y_{ik}  \mbox{\ \ for all\ \ } v_i \in S,\ k \in \{1, 2, \ldots, m\}
\end{split}
\end{equation}
whose solution is  
\begin{equation}\label{eq:solutionST}
\bF^*_U = (\bD_{UU} -\bW_{UU})^{-1}\bW_{US}\hspace{0.05cm} \bF^*_S	
\end{equation}
where $\bF^*_S = \bY_S$.
\fi

When learning multiple related tasks, a widely used approach is requiring that similar tasks be assigned similar labelings. To this end, we introduce the linear map $\psi_{\bscA^{-1}}:\mathbb{R}^{n\times m} \to \mathbb{R}^{n\times m}$, defined as follows: 
\begin{equation}\label{eq:invmap}
    \psi_{\bscA^{-1}} (\bY) = \bY \bscA^{-1}~.
\end{equation}
It can be shown that the map $\psi_{\bscA^{-1}}$ acts on a multitask labeling matrix $\bY$ by getting closer (in Euclidean distance) the labelings (columns of $\bY$) corresponding to tasks that are similar according to $\bscC$.

By means of $\psi_{\bscA^{-1}}$, the exploitation of task similarities can be encoded into the learning problem~(\ref{eq:objective}) as follows: 
\begin{equation}
\label{eq:MT_invlabprop}
\begin{array}{l}
%	&\min_{\bbf\in \{-1,1 \}^{n}}\|\bbf\|_L^2\\
{\displaystyle \min_{\bF} \trace (\bF^{\top}\bL\bF) }
\\
\text{s.t.} \quad F_{ik} = \wY_{ik}  \quad i \in S,\, k=1,\dots,m
\end{array}
\end{equation}
where $\wbY = \psi_{\bscA^{-1}}(\bY) = \bY \bscA^{-1}$. The solution to~(\ref{eq:MT_invlabprop}) is   
\begin{displaymath}\label{eq:solutionMT}
\widetilde\bF_U = (\bD_{UU} - \bW_{UU})^{-1}\bW_{US} \wbY_S	
\end{displaymath}
where $\widetilde{\bF}_U$ is the submatrix of $\bF$ including only the rows indexed by $U$, and $\wbY_S$ is the submatrix of $\wbY$ including only the rows indexed by $S$. By observing that $\wbY_S = \bY_S \bscA^{-1}$, we have
\[
	\widetilde\bF_U = \big(\bD_{UU} -\bW_{UU})^{-1}\bW_{US} \bY_S \bscA^{-1} =  \bF_U^*\bscA^{-1}
\]
where $\bF_U^*$ is the solution of~(\ref{eq:MT_invlabprop}) with constraints $F_{ik} = Y_{ik}$ for $i \in S$ and $k=1,\dots,m$. The equality $\widetilde\bF_U = \bF_U^*\bscA^{-1}$ shows that it is equivalent to apply the task feature map~(\ref{eq:invmap}) before or after performing label propagation. This ensures that the multitask mapping does not increase the label propagation complexity. 

As we show next, this solution does not perform well on unbalanced classification problems, where some class (typically the positive class) is largely underrepresented. 
We propose here an alternative approach, which exploits the prior information about task relatedness in an ``inverse'' manner. Specifically, we propose a multitask label propagation algorithm which learns multiple tasks by requiring that dissimilar tasks be assigned dissimilar labelings. As we see in the experiments, this approach turns out to work particularly well on unbalanced classification problems. 

The first component of this method is a dissimilarity matrix $\bar\bscC$, where $\bar\scC_{kr} \in [0,1]$ is measure of dissimilarity between tasks $k$ and $r$ (we discuss in Section~\ref{subsub:simmat} possible choices for the matrices $\bscC$ and $\bar\bscC$).  

Given the matrix $\bar\bscC$, we consider the multitask map $\psi_{\bar\bscA} : \mathbb{R}^{n\times m} \to \mathbb{R}^{n\times m}$, defined as
\begin{equation}\label{eq:map_over_A}
 \psi_{\bar\bscA} (\bY) = \bY \bar\bscA
\end{equation}
where $\bar\bscA = \bar\gamma \bscI_m + \bar\bscL$, $\bar\gamma > 0$, and $\bar\bscL$ is the Laplacian matrix of $\bar\bscC$. Unlike the inverse transformation~(\ref{eq:invmap}), the map $\psi_{\bar\bscA}$ moves the columns of matrix $\bY$ farther away from each other in the corresponding $n$-dimensional space (in the sense of the Euclidean distance). We formally show that in Section~\ref{subsubsec:intMatProp}. Using $\psi_{\bar\bscA}$ instead of $\psi_{\bscA^{-1}}$ in~(\ref{eq:MT_invlabprop}), we obtain the following optimization problem:
\begin{equation}\label{eq:MT_labprop}
	\begin{array}{l}
%	&\min_{\bbf\in \{-1,1 \}^{n}}\|\bbf\|_L^2\\
	{\displaystyle \min_{\bF} \trace \big(\bF^{\top}\bL\bF\big)}
\\
	\text{s.t.} \quad F_{ik} = \widehat Y_{ik}  \quad i \in S,\, k=1,\dots,m
\end{array}
\end{equation}
with $\widehat \bY = \psi_{\bar\bscA}(\bY)$. Similarly to~(\ref{eq:MT_invlabprop}), the solution of~(\ref{eq:MT_labprop}) is
\[
	\widehat\bF_U = \big(\bD_{UU} -\bW_{UU})^{-1}\bW_{US} \bY_S \bar{\bscA} =  \bF_U^*\bar{\bscA}
\]
where $\bF_U^*$ is the solution of~(\ref{eq:MT_labprop}) with constraints $F_{ik} = Y_{ik}$ for $i \in S$ and $k=1,\dots,m$. Just like in the previous case, the equality $\widetilde\bF_U = \bF_U^*\bar{\bscA}$ shows that it is equivalent to apply the task feature map~(\ref{eq:map_over_A}) before or after performing label propagation.

We call MTLP-inv the similarity-based method~(\ref{eq:MT_invlabprop}) and MTLP the dissimilarity-based method~(\ref{eq:MT_labprop}). In the next section we show some interesting properties of the map $\psi_{\bar\bscA}$ which make MTLP suitable for unbalanced classification problems.

%%%%%%%%%%%%%%%%%%%%%%%%%%%%%%%%%%%%%%%%%%%%%%%%%%%%%%%%%%%%%%%%%%%%%%%%%%%%%%%%%%%%%%%%%%%%5
\subsubsection{Analysis of the multitask map $\psi_{\bar\bscA}$}\label{subsubsec:intMatProp}
Given $\bM \in \mathbb{R}^{n\times m}$, let $\bM_{i\cdot}$ and $\bM_{\cdot k}$ be, respectively, the $i$-th row and the $k$-th column of the matrix $\bM$. Let also $\mathcal{P}_i = \{ 1 \le k \le m \,:\, Y_{ik} = 1\}$ be the set of tasks for which the instance $i$ is positive, and $\mathcal{N}_i$ the set of tasks for which the instance $i$ is negative. 
We introduce the following notation: for each $k=1,\dots,m$
\[
	\mathfrak{d}^+_{k,i} = \sum_{r\in \mathcal{P}_i} \bar\scC_{rk}
\qquad
	\mathfrak{d}^-_{k,i} = \sum_{r\in \mathcal{N}_i} \bar\scC_{rk}
\qquad
	\mathfrak{d}_k = \sum_{r=1}^m \bar\scC_{rk}
\]
and
\[
	\mathfrak{a}^+_{k,i} = \sum_{r\in \mathcal{P}_i} \bar \scA_{rk}
\qquad
	\mathfrak{a}^-_{k,i} = \sum_{r\in \mathcal{N}_i} \bar A_{rk}
\qquad
	\mathfrak{a}_k = \sum_{r=1}^m \bar\scA_{rk}~.
\]
The next result shows that the action of the linear map $\psi_{\bar\bscA}$ on the label matrix $\bY$ is to change the value of each label without altering the sign. The label of an instance $i$ in task $k$ is made roughly proportional to the weighted sum of tasks in $\bar\bscC$ that are dissimilar to $k$ and have a different label for instance $i$ ---see also Corollary~\ref{cor:1}.
%%%%%%%%%%%%%%%%%%%%%%%%%%%%%%%%%%%%%%%%%%%%%%%%%%%%
 \begin{theorem}{}\label{th:1}
Given $\bY \in \{-1, 1\}^{n\times m}$, the task interaction matrix $\bar\bscC \in \mathbb{R}^{m\times m}$, and the map $\psi_{\bar\bscA}: \mathbb{R}^{n\times m} \longrightarrow \mathbb{R}^{n\times m}$ such that $\widehat\bY=\psi_{\bar\bscA} (\bY) = \bY \bar\bscA$, where $\bar\bscA = \bar\gamma \bscI_m + \bar\bscL$, then for all $i=1,\dots,n$ it holds
\begin{displaymath}
\widehat Y_{ik} = \left\{ \begin{array}{ll}
\ \ \bar\gamma + 2\mathfrak{d}^-_{k,i}& \mbox{\hspace{1cm}  if \ } Y_{ik} = +1 \\
-\bar\gamma -2\mathfrak{d}^+_{k,i} & \mbox{\hspace{1cm}  if \ } Y_{ik} = -1 \\
\end{array}
\right.
\end{displaymath}
%%%%%%%%%%%%%%%%%%%%%%%%%%%%%%%%%%%%%%%%%%%%%%%%%%%%%%%%% 
\label{th:hatY_prop}\end{theorem}
\begin{IEEEproof}
By definition, $\widehat Y_{ik} = \sum_{r=1}^m Y_{ir}\bar\scA_{rk} = \mathfrak{a}^+_{k,i} - \mathfrak{a}^-_{k,i}$.
We distinguish the following two cases.

\smallskip\noindent\textbf{Case 1.}
$k \in \mathcal{P}_i$. In this case we have $\mathfrak{a}^+_{k,i} = \bar\scA_{kk} - \mathfrak{d}^+_{k,i} = \bar\gamma + \mathfrak{d}_k - \mathfrak{d}^+_{k,i} = \bar\gamma + \mathfrak{d}^+_{k,i}+\mathfrak{d}^-_{k,i} - \mathfrak{d}^+_{k,i} = \bar\gamma + \mathfrak{d}^-_{k,i}$, since by definition $\mathfrak{d}_k = \mathfrak{d}^+_{k,i}+\mathfrak{d}^-_{k,i}$ for any $i\in \{1,2,\ldots,n\}$. Moreover, since $k \in \mathcal{P}_i$, we have $ \mathfrak{a}^-_{k,i}\ =\ -\mathfrak{d}^-_{k,i}$ (by the definition of Laplacian), and accordingly 
\begin{displaymath}
\widehat Y_{ik}\ =\ \bar\gamma +\mathfrak{d}^-_{k,i}- (-\mathfrak{d}^-_{k,i})  = \bar\gamma + 2\mathfrak{d}^-_{k,i}
\end{displaymath}
\textbf{Case 2.} $k \in \mathcal{N}_i$. In this case, it holds  $\mathfrak{a}^+_{k,i} = -\mathfrak{d}^+_{k,i}$, whereas $\mathfrak{a}^-_{k,i} = \bar\scA_{kk} -\mathfrak{d}^-_{k,i} = \bar\gamma + \mathfrak{d}_k -\mathfrak{d}^-_{k,i} = \bar\gamma + \mathfrak{d}^+_{k,i}$. It follows
\begin{displaymath}
\widehat Y_{ik}\ =\ -\mathfrak{d}^+_{k,i} - \bar\gamma - \mathfrak{d}^+_{k,i}= - \bar\gamma - 2\mathfrak{d}^+_{k,i}
\end{displaymath}
The property is proven by observing that $k \in \mathcal{P}_i$ implies $Y_{ik} = +1$ and $k \in \mathcal{N}_i$ implies $Y_{ik} = -1$. 
\end{IEEEproof}
Using Fact~\ref{th:1} we can show that the map $\psi_{\bar\bscA}$ tends to increase the distance between the labelings $\bY_{\cdot r}$ and $\bY_{\cdot s}$, for any pair of distinct tasks $r, s \in \{1, 2, \ldots, m\}$. Indeed, we can prove the following.
 \begin{theorem}{} 
Given $\bY \in \{-1, 1\}^{n\times m}$, the task interaction matrix $\bar\bscC \in \mathbb{R}^{m\times m}$, and the map $\psi_{\bar\bscA}: \mathbb{R}^{n\times m} \longrightarrow \mathbb{R}^{n\times m}$ such that $\widehat\bY=\psi_{\bar\bscA} (\bY) = \bY \bar\bscA$, where $\bar\bscA = \bar\gamma \bscI_m + \bar\bscL$. Then for every $r,s \in \{1, 2, \ldots, m\}$ it holds
\begin{displaymath}
\|\bY_{\cdot r} \ -\ \bY_{\cdot s}\|^2 \leq \|\widehat\bY_{\cdot r}\ -\ \widehat\bY_{\cdot s}\|^2
\end{displaymath}
for every $\bar\gamma \geq 1$, where $\| \cdot \|$ is the Euclidean norm.
\label{th:A_prop}\end{theorem}
\begin{IEEEproof}
We prove this property by showing that $(Y_{ir} - Y_{is})^2 \leq (\widehat Y_{ir} - \widehat Y_{is})^2$ for all $i\in\{1, 2, \ldots, n\}$. We distinguish the following four cases:
 \begin{enumerate}[\textbf{Case} 1:]
\item $Y_{ir}=Y_{is}=1$. In this case $(Y_{ir}-Y_{is})^2\ =\ 0$, and by Fact~\ref{th:1}, $(\widehat Y_{ir}-\widehat Y_{is})^2\ =\ (\bar\gamma +2\mathfrak{d}^-_{r,i} -\bar\gamma -2\mathfrak{d}^-_{s,i} )^2\ =\ 4(\mathfrak{d}^-_{r,i} - \mathfrak{d}^-_{s,i})^2 \geq 0$. 
%Since $Y_{ir}=Y_{is}=1$, we have $0\leq |\mathcal{N}_i|\leq m-2$, which means $\mathfrak{d}^-_{r,i},\mathfrak{d}^-_{s,i} \geq 0$. Accordingly $(\mathfrak{d}^-_{r,i} - \mathfrak{d}^-_{s,i})^2 \geq 0$, and hence $(Y_{ir} - Y_{is})^2 \leq (\hat Y_{ir} - \hat Y_{is})^2$.
\item $Y_{ir}=Y_{is}=-1$. Even in this case $(Y_{ir}-Y_{is})^2 = 0$, whereas $(\widehat Y_{ir}-\widehat Y_{is})^2 = (-\bar\gamma -2\mathfrak{d}^+_{r,i} + \bar\gamma +2\mathfrak{d}^+_{s,i})^2\ =\ 4(\mathfrak{d}^+_{s,i} - \mathfrak{d}^+_{r,i})^2 \geq 0$. 
%Similarly to the previous case, having that $0\leq |\mathcal{P}_i|\leq m-2$, it follows that $(\mathfrak{d}^+_{s,i} - \mathfrak{d}^+_{r,i})^2 \geq 0$, and hence $(Y_{ir} - Y_{is})^2 \leq (\hat Y_{ir} - \hat Y_{is})^2$.
\item $Y_{ir}= 1\ \wedge\ Y_{is}=-1$. In this case, $(Y_{ir}-Y_{is})^2 = 4$, and $(\widehat Y_{ir}-\widehat Y_{is})^2 = (\bar\gamma +2\mathfrak{d}^-_{r,i} + \bar\gamma +2\mathfrak{d}^+_{s,i})^2\ =\ 4(\bar\gamma + \mathfrak{d}^-_{r,i} + \mathfrak{d}^+_{s,i})^2$. Since both $\mathfrak{d}^+_{s,i}, \mathfrak{d}^-_{r,i} \geq 0$ and $\bar\gamma \geq 1$, it follows $(Y_{ir} - Y_{is})^2 \leq (\widehat Y_{ir} - \widehat Y_{is})^2$.
%\begin{displaymath}
%\begin{split}
%Y_{ir} = 1 \wedge\ Y_{is}=-1\ &\Longrightarrow\ 1\leq  |\mathcal{P}_i|, |\mathcal{N}_i|\leq m-1\ \\&\Longrightarrow\ \mathfrak{d}^+_{s,i}, \mathfrak{d}^-_{r,i} \geq 0,
%\end{split}
%\end{displaymath}
%, since .
\item $Y_{ir}= -1\ \wedge\ Y_{is}=1$. Again $(Y_{ir}-Y_{is})^2 = 4$, and $(\widehat Y_{ir}-\widehat Y_{is})^2 = (-\bar\gamma -2\mathfrak{d}^+_{r,i} - \bar\gamma -2\mathfrak{d}^-_{s,i})^2\ =\ 4(-(\bar\gamma + \mathfrak{d}^+_{r,i} + \mathfrak{d}^-_{s,i}))^2$. As $\mathfrak{d}^-_{s,i}, \mathfrak{d}^+_{r,i} \geq 0$ and $\bar\gamma \geq 1$
%\begin{displaymath}
%\begin{split}
%Y_{ir}= -1 \wedge\ Y_{is}=1\ &\Longrightarrow\ 1\leq  |\mathcal{P}_i|, |\mathcal{N}_i|\leq m-1\ \\&\Longrightarrow\ \mathfrak{d}^+_{r,i}, \mathfrak{d}^-_{s,i} \geq 0,
%\end{split}
%\end{displaymath}
we have, like the previous case, $(Y_{ir} - Y_{is})^2 \leq (\widehat Y_{ir} - \widehat Y_{is})^2$.
\end{enumerate}
\end{IEEEproof}
%Fact~\ref{th:A_prop} thereby shows that the applied multi-task transformation allows to achieve our initial aim of learning multiple tasks by assigning to dissimilar tasks more distant labelings.
%
%NCB Added
The map $\psi_{\bar\bscA}$ not only increases the distance between the instance-indexed label vector for two distinct tasks (as we just showed), but it also increases the distance between the task-indexed label vector for two distinct instances. Indeed, since $\bar\bscL$ is positive semidefinite, it is easy to show that when $\bar{\gamma} \ge 1$ the transformation $\psi_{\bar\bscA}$ increases the distance between the labelings $\bY_{i \cdot}$ and $\bY_{j \cdot}$, for any pair of distinct instances $i,j \in \{1, 2, \ldots, n\}$.

We now focus our discussion on another important feature of the algorithm, which makes our multitask label propagation appropriate for tasks with very unbalanced labelings. Specifically, when most entries of each column in the label matrix $\bY$ are $-1$. In this case, the rows of $\bY$ also contain mostly negative entries. Accordingly, by Fact~\ref{th:1}, we can compensate the preponderance of negatives by applying the map $\psi_{\bar\bscA}$. We show that with an example.

\begin{figure}[!t]
\begin{center}
\scriptsize
\begin{tabular}{c}
\hspace{-0.5cm} \includegraphics [width=9.2cm]  {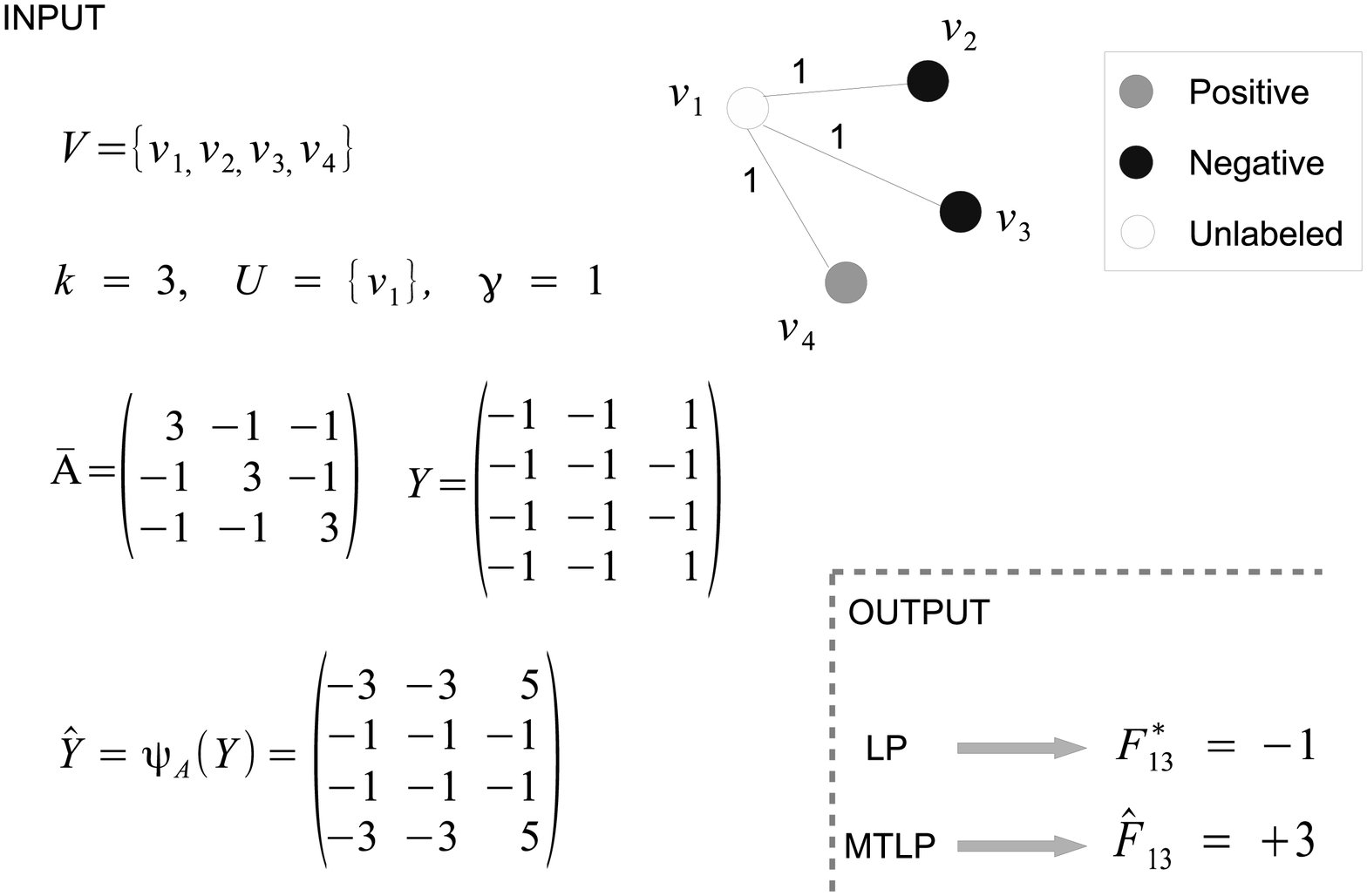}	 
\end{tabular}
\vspace{-1cm}
\caption{Toy example with four vertices $v_1, \ldots v_4$, labeled for three  tasks according to the matrix $\bY$. The test point is instance $v_1$ in all the tasks, and we apply LP and MTLP to predict it. For tasks $1$ and $2$ both methods correctly associate $v_1$ with a negative label. However, in the third task, only MTLP correctly predicts a positive label for $v_1$.}\label{fig:toyMTLPexample}
\normalsize
\end{center}
\end{figure}

Consider the task interaction matrix $\bar\bscC$ such that $\scC_{rs}=1$ for all $r \neq s$. That is, all tasks are strongly dissimilar to each other. Then
\begin{equation}\label{eq:A1}
 \bar\bscA = \left[ \begin{array}{cccc}
\bar\gamma +m-1 & -1 & \ldots & -1 \\
-1 & \bar\gamma +m-1 & \ldots & -1 \\
\vdots & \vdots & \vdots & \vdots \\
-1 & \ldots & \ldots & \bar\gamma +m-1 \\
 \end{array}
\right]
\end{equation}
By Fact~\ref{th:1}, it is straightforward to prove the following.
 \begin{corollary}{}\label{cor:1}
Fix $\bY \in \{-1, 1\}^{n\times m}$ and the map $\psi_{\bar\bscA}: \mathbb{R}^{n\times m} \to \mathbb{R}^{n\times m}$ such that $\hat\bY=\psi_{\bar\bscA} (\bY) = \bY \bar\bscA$, where $\bar\bscA$ is defined as in~(\ref{eq:A1}). Then, for all $i=1,\ldots,n$ it holds that
\begin{displaymath}
\widehat Y_{ik} = \left\{ \begin{array}{ll}
\ \ \bar\gamma  +2|\mathcal{N}_i|& \mbox{\hspace{1cm}  if \ } Y_{ik} = +1 \\
-\bar\gamma -2|\mathcal{P}_i| & \mbox{\hspace{1cm}  if \ } Y_{ik} = -1. \\
\end{array}
\right.
\end{displaymath}   
\end{corollary}
Corollary~\ref{cor:1} shows that, when $|\mathcal{P}_i| \ll |\mathcal{N}_i|\ =\ m-|\mathcal{P}_i|$ 
(that is, the multitask labeling for vertex $i$ is unbalanced towards negatives), the map $\psi_{\bar\bscA}$ assigns to positives ($Y_{ik} = +1$) an absolute value higher than that assigned to negatives ($Y_{ik} = -1$). An analogous behaviour characterizes our method when a generic matrix $\bar\bscC$ is considered, as stated in Fact~\ref{th:1}.
% since when $|\mathcal{P}_i| < |\mathcal{N}_i|$, it holds $\mathfrak{d}^-_{k,i} > \mathfrak{d}^+_{k,i}$, for the majority of $k\in \{1,2,\ldots,m\}$. 
This simple property allows the rare positive labels to propagate in the graph. This is unlike the standard LP algorithm, where positive vertices are easily overwhelmed by the negative vertices during the label propagation process. The toy example in Figure~\ref{fig:toyMTLPexample} shows that the application of the map $\psi_{\bar\bscA}$, where $\bar\bscA$ is defined as in~(\ref{eq:A1}), allows to improve the final classification of vertices. These observations are empirically confirmed in Section~\ref{sec:exp}. 
%it is expected $|\mathcal{P}_i| < |\mathcal{N}_i|$ for the large majority of $i \in \{1,2 , \ldots, n\}$. 

\subsubsection{Task similarities}\label{subsub:simmat}\label{subsub:simMat}
While MTLP and MTLP-inv are designed to work with any task matrix, similarity and dissimilarity measures are typically tailored to specific domains. Different tasks may share different types of similarities, or may be organized in a hierarchy with a specific structure ---such as a tree or a directed acyclic graph--- where the positive instances of the children tasks are subsets of the positive instances of their parent tasks. In the case of a hierarchy, different approaches for computing the task matrix are possible: considering only the structure of the hierarchy~\cite{Wu94,Leacock98}, or combining the hierarchical information with the information content of the tasks~\cite{Meng13}.
%In particular, in this section we consider two similarity measures and their corresponding dissimilarities from the second category: the Resnick's~\cite{Resnik} and Jiang's~\cite{Jiang} measures.

In this work we consider two dissimilarity measures ($\diss_0$ and $\diss_3$) and three similarity measures ($\simm_1,\simm_2,\simm_3$). The similarity measures $\simm_1$ and $\simm_2$ were introduced by Jiang~\cite{Jiang97} and Lin~\cite{Lin98}, respectively. Both measures are derived from the dissimilarity measure $\diss_0$, whose definition requires a hierarchy over the tasks. The dissimilarity $\diss_3$ is computed directly from the similarity $\simm_3$, which does not require any hierarchical information.

When tasks are organized in a hierarchy, we denote by $\anc(k) \subset \{1,\ldots,m\}$ the set of ancestor tasks of task $k$ in the hierarchy. Moreover, we use $\nu(k)$ to denote the frequency of positive instances for task $k$. Since a positive instance for a task $k$ is also positive for any $r\in \anc(k)$, it holds that $\nu(k)\leq \nu(r)$. Finally, we denote by $\MA(k,r)$ the common ancestor of tasks $k$ and $r$ whose frequency $\nu(\MA(k,r))$ is the lowest among all ancestors of $k$ and $r$. 

%The Resnick's measure between two concepts $k$ and $r$ is
%\begin{displaymath}
%sim_{R}(k,r) = -\log\p(\MA(k,r))
%\end{displaymath}
%where $-log(\p(\MA(k,r)))$ is the information content of the minimum common ancestor of $k$ and $r$. A map $\phi_R$ which allows to compute the dissimilarity of two concept $k$ and $r$ from their similarity is
%\begin{displaymath}
%\phi_R(k,r) = \frac{1}{1+}
%\end{displaymath}
Let $-\log(\nu(k))$ be the information content of task $k$. We start by recalling the hierarchical dissimilarity measure introduced in~\cite{Lin98},
\[
	\diss_0(k, r) = -\log \nu(k) - \log\nu(r) + 2\log\nu(\MA(k,r))~.
\]
This is the sum of the information content of $k$ and $r$ minus the information content of their closest common ancestor $\MA(k,r)$. Note that $\diss_0$ is always positive, as $\MA(k,r) \ge \max\{\nu(k),\nu(r)\}$. The two hierarchical similarity measures associated with $\diss_0$ are defined as follows.

\noindent\textit{Jiang similarity measure:}
\begin{displaymath}
\simm_1(k,r) = \frac{1}{1+\diss_0(k,r)}~.
\end{displaymath} 
\textit{Lin similarity measure:}
\begin{displaymath}
\simm_2(k,r) = \frac{2\log\nu(\MA(k,r))}{\log \nu(k) + \log \nu(r)}
\end{displaymath}
Our third similarity measure does not rely on a hierarchy of tasks. Let $P^{(k)}$ the set of instances that are positive for the task $k$.

\noindent\textit{Information content measure:}
\begin{displaymath}
\simm_3(k,r) =  \left\{ \begin{array}{cl}
{\displaystyle \frac{\big|P^{(k)}\cap P^{(r)}\big|}{\big|P^{(k)}\cup P^{(r)}\big|} } & \text{if $P^{(k)}\cup P^{(r)} \neq \emptyset$} \\
\\
0  & \text{otherwise.}
\end{array}
\right.
\end{displaymath}
This is the ratio between the number of examples that are positive for both tasks and the number of examples that are positive for at least one task. The higher the number of shared positive examples, the higher the similarity (up to $1$). When two tasks do not share any positive example, their similarity is zero. In a hierarchy of tasks, tasks with many positive examples are usually closer to the root (less specific). In this case the denominator of $\simm_3$ tends to reduce the similarity between the two tasks as opposed to the case in which tasks have a small number of positive annotations. Indeed, sharing annotations between two specific tasks (closer to leaves) is more informative than sharing annotations between two more general tasks (closer to the root).

In the experiments, we compare learning with similarities $\simm_1(k,r)$ and $\simm_2(k,r)$ against learning with the dissimilarity $\diss_0(k,r)$. We also compare learning with $\simm_3(k,r)$ against $\diss_3(k,r) = 1 - \simm_3(k,r)$. For each one of the similarity/dissimilarity measures defined above, we set $\scC_{kr} = \simm(k,r)$ and $\bar\scC_{kr} = \diss(k,r)$ (where necessary, values are normalized so that all matrix entries lie in the range $[0,1]$).

\section{Results and Discussion}\label{sec:exp}
In this section we evaluate our multitask algorithms on the prediction of the bio-molecular functions of proteins belonging to some considered model organisms. We start by describing the experimental setting. Then we compare the performance of our algorithms against that of state-of-the-art methods.
\subsection{Experimental setting}\label{sub:exp_sett}
\subsubsection{Data}
We considered three different experiments to predict the protein functions of three model organisms: {\em Drosophila melanogaster} (fly), {\em Homo sapiens} (human) and {\em Escherichia coli} (bacteria). Gene networks for model organisms have been downloaded from the GeneMANIA website~({\tt www.genemania.org}), and selected in order to cover different types of data, including co-expression, genetic interactions, shared domains, and physical interactions. The selected networks are described in Tables~\ref{tab:fly_data},~\ref{tab:human_data} and ~\ref{tab:bacteria_data}.
%%%%%%%%%%%%%%%%%%%%%%%%%%%%%%%%%%
\begin{table}[h!]
\scriptsize
\begin{center}
\begin{tabular}{@{}lcc@{}}
\hline\\[-6pt] 
\bf{Type} & {\bf{Source}} & {\bf{Nodes}}\\[0.5pt] 
\hline\\[-3pt]
Co-expression & Baradaran-Heravi et al.~\cite{Baradaran12} & 8857\\[2pt]
Co-expression & Busser et al.~\cite{Busser12} & 8857\\[2pt]
Co-expression & Colombani et al.~\cite{Colombani12} & 8857\\[2pt]
Co-expression & Lundberg et al.~\cite{Lundberg12} & 8857\\[2pt]
Genetic interactions & BioGRID~\cite{Biogrid06} & 929\\[2pt]
Genetic interactions & Yu et al.~\cite{Yu08} & 1414\\[2pt]
Physical interactions & Guruharsha et al. A~\cite{Guruharsha11} & 1866\\[2pt]
Physical interactions & Guruharsha et al. B~\cite{Guruharsha11} & 3833\\[2pt]
Physical interactions & BioGRID~\cite{Biogrid06} & 558\\[2pt]
Shared protein domains & InterPro~\cite{InterPRO} & 5627\\[2pt]
\hline\\[-4pt]
\end{tabular} 
\end{center}
\caption{Fly networks.}
\normalsize
\label{tab:fly_data}
\end{table}
%%%%%%%%%%
%%%%%%%%%%%%%%%%%%%%%%%%%%%%%%%%%%
\begin{table}[h!]
\scriptsize
\begin{center}
\begin{tabular}{@{}lcc@{}}
\hline\\[-6pt] 
\bf{Type} & {\bf{Source}} & {\bf{Nodes}}\\[0.5pt] 
\hline\\[-3pt]
Co-expression & Bahr et al.~\cite{Bahr13} & 7611\\[2pt]
Co-expression & Balgobind et al.~\cite{Balgobind11} & 17522\\[2pt]
Co-expression & Bigler et al.~\cite{Bigler13} & 17522\\[2pt]
Co-expression & Botling et al.~\cite{Botling13} & 17522\\[2pt]
Co-expression & Clarke et al.~\cite{Clarke13} & 17458\\[2pt]
Co-expression & Vallat et al.~\cite{Vallat13} & 17521\\[2pt]
Common biological & PATHWAYCOMMONS~\cite{Cerami11} & 2133\\[2pt]
\hspace{0.1cm}pathways && \\[2pt]
Common biological & Wu et al.~\cite{Wu10} & 5319\\[2pt]
\hspace{0.1cm}pathways && \\[2pt]
Physical interactions & BioGRID~\cite{Biogrid06} & 15800\\[2pt]
Physical interactions & iRref-GRID~\cite{Razick08} & 9403\\[2pt]
Physical interactions & iRref-HPRD~\cite{Razick08} & 9403\\[2pt]
Physical interactions & iRref-OPHID~\cite{Razick08} & 9403\\[2pt]
Physical interactions & IREF SMALL-SCALE-STUDIES~\cite{Razick08} & 9036\\[2pt]
Shared protein  & InterPro~\cite{InterPRO} & 15800\\[2pt]
\hspace{0.1cm}domains && \\[2pt]
Shared protein  & Pfam~\cite{Finn16} & 15251\\[2pt]
\hspace{0.1cm}domains && \\[2pt]
\hline\\[-4pt]
\end{tabular} 
\end{center}
\caption{Human networks.}
\normalsize
\label{tab:human_data}
\end{table}
%%%%%%%%%%
%%%%%%%%%%%%%%%%%%%%%%%%%%%%%%%%%%
\begin{table}[h!]
\scriptsize
\begin{center}
\begin{tabular}{@{}lcc@{}}
\hline\\[-6pt] 
\bf{Type} & {\bf{Source}} & {\bf{Nodes}}\\[0.5pt] 
\hline\\[-3pt]
Co-expression & Graham et al.~\cite{Graham12} & 3959\\[2pt]
Co-expression & Robbins-Manke et al.~\cite{Robbins05} & 3912\\[2pt]
Genetic interactions & Babu et al.~\cite{Babu11} & 715\\[2pt]
Genetic interactions & Butland et al.~\cite{Butland08} & 3497\\[2pt]
Physical interactions &  Hu at al \cite{Hu09} & 1537\\[2pt]
Physical interactions & IREF-Dip~\cite{Razick08} & 633\\[2pt]
Physical interactions & Y2H - PPI & 1063\\[2pt]
Shared protein domains & InterPro~\cite{InterPRO} & 3005\\[2pt]
Shared protein domains & Pfam~\cite{Finn16} & 2726\\[2pt]
\hline\\[-4pt]
\end{tabular} 
\end{center}
\caption{Bacteria networks.}
\normalsize
\label{tab:bacteria_data}
\end{table}
%%%%%%%%%%
%%%%%%%%%%%%%%%%%%%%%%%%%%%%%%%%%%%%%%%%%%%%%%%%%%%%%%%%%5
For every organism, networks were integrated through unweighted sum on the union of genes in the individual networks. No preprocessing was applied to the individual networks, whereas each network, denoted by the corresponding connection matrix $\bW$, was normalized as follows:
\begin{displaymath}\label{eq:normaliz}
\hat{\bW} = {\bD}^{-1/2}  \bW \bD^{-1/2}
\end{displaymath} 
where $\bD$ is the diagonal matrix with diagonal entries $d_{ii} = \sum_j W_{ij}$.

%No preprocessing has been applied to single networks, since GeneMANIA networks already provide a real score for each pair of genes representing a measure of their functional similarity.

Protein functions were downloaded from the Gene Ontology. This ontology is structured as a directed acyclic graph with different levels of specificity and contains three branches: \textit{Biological Process} (BP), \textit{Molecular Functions} (MF), and \textit{Cellular Components} (CC). We considered the experimental annotations in the releases 07.03.16, 16.03.16, and 17.10.16 respectively for \textit{fly}, \textit{human} and \textit{bacteria} organisms. We performed a dedicated experiment for every branch.
 
For predicting the most specific terms in the ontology (i.e., those best describing protein functions), and in order to consider terms with a minimum amount of prior information, we selected all the GO terms with $5-100$ positive annotated genes, obtaining $2657$ ($1742$ BP, $539$ MF, $376$ CC), $5312$ ($3799$ BP, $957$ MF, $556$ CC), and $1324$ ($653$ BP, $610$ MF, $61$ CC) terms for \textit{fly}, \textit{human}, and \textit{bacteria}, respectively. We considered two groups of GO terms according to their specificity: GO terms with $5$-$20$ and $21$-$100$ annotated proteins, for a total of $2$ categories for every GO branch.
In the end, we obtained a total of $10329$ \textit{fly}, $15262$ \textit{human}, and $4132$ \textit{bacteria} genes which have at least one GO positive annotation in the considered GO release. The obtained tasks are therefore severely unbalanced toward negatives.
%Finally, the pairwise task interaction matrix $\bC$ has been obtained by adopting the \textit{GOSSTO} tool~\cite{Vale14d}. 
%and 3 GO networks (release 23-3-13 for yeast and 15-5-13 for fly),
\subsubsection{Evaluation metrics}
In order to evaluate the generalization performance of the compared methods, we applied a $3$-fold cross-validation experimental setting and adopted the Area Under the Precision-Recall Curve (AUPRC) as ``per term'' ranking measure. AUPRC is indeed more informative on unbalanced settings than the classical area under the ROC curve~\cite{Saito15}.
Furthermore, following the recent CAFA2 international challenge, we also considered a ``protein-centric evaluation'' to assess performance accuracy in predicting all ontological terms associated with a given protein sequence~\cite{CAFA2}. In this scenario, the multiple-label F-score is used as performance measure. More precisely, if we indicate as $\TP_j(t)$, $\TN_j(t)$ and $\FP_j(t)$ respectively the number of true positives, true negatives, and false positives for the protein $j$ at threshold $t$, we can define the ``per-protein'' multiple-label precision $\Prec(t)$ and recall $\Rec(t)$ at a given threshold $t$ as:
\begin{align*}
\Prec(t) &= \frac{1}{n} \sum_{j=1}^{n} \frac{\TP_j(t)}{\TP_j(t) + \FP_j(t)}
\\
\Rec(t) &= \frac{1}{n} \sum_{j=1}^{n} \frac{\TP_j(t)}{\TP_j(t) + \FN_j(t)}
\label{eq:prec-rec}
\end{align*}
where $n$ is the number of proteins. In other words, $\Prec(t)$ (resp., $\Rec(t)$) is the average multilabel precision (resp., recall) across proteins.
The multilabel F-measure depends on $t$ and according to CAFA2 experimental setting, the maximum achievable F-score ($\Fmax$) is adopted as the main multilabel ``per-protein'' metric:
\begin{equation}
\Fmax = \max_t \frac{2 \Prec(t) \Rec(t)}{\Prec(t) + \Rec(t)} 
\label{eq:F}
\end{equation}
%Traditional methods, where particular genes of interest are investigated based solely on positional information, are not well suited for most common genetic disorders, such as autism, due to complex gene patterns, such as gene epistasis.  Disease gene prioritization addresses this problem by generating ordered lists of candidates pertaining a particular gene-disease relatedness score. To overcome the limitations of traditional methods,
\subsection{Results}
\subsubsection{Evaluating GO semantic similarities}
This section investigates the impact of the task similarity/dissimilarity measures described in Section~\ref{subsub:simMat} on the performance of the proposed multitask label propagation algorithms.
Table~\ref{tab:simmeas_comp} shows the obtained results. In this experiment we set $\gamma = \bar\gamma =1$ (the choice of parameter $\bar\gamma$ is discussed in Section~\ref{subsub:gamma}). When MTLP-inv uses the similarity measures $\simm_1,\simm_2$ and MTLP uses $\diss_0$, MTLP outperforms MTLP-inv in both AUPRC and $\Fmax$. Nevertheless, the GO term similarity $\simm_3$ is much more informative for MTLP-inv, which achieves in this case results competitive with MTLP (whose performance instead is nearly indistinguishable when using $\diss_0$ or $\diss_3$), and in some cases even better. The difference in favor of MTLP seems to increase with the data imbalance: on \textit{human} data set, the most unbalanced, we observe the highest gap in favor of MTLP; whereas on the \textit{Bacteria} data set, the least unbalanced, the gap is reduced and ---in some cases like for the MF terms--- MTLP-inv significantly outperforms MTLP in terms of average AUPRC. In terms of $\Fmax$, however, MTLP is always the top method.
 
\begin{figure}[h!]
\begin{center}
\scriptsize
\begin{tabular}{c}
\hspace{-0.5cm} \includegraphics [angle=-90, width=0.5\textwidth]  {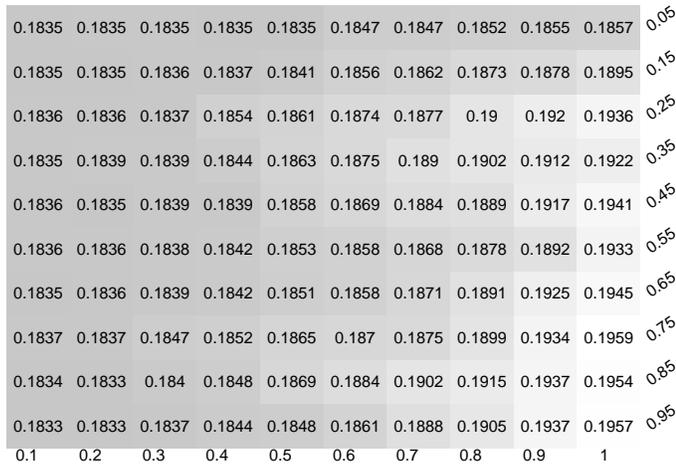}	 
\end{tabular}
\caption{Average AUPRC values achieved by MTLP method of \textit{fly} data and CC GO terms when the matrix $\bar\bscC$ is randomly generated. Values of $\tau$ are reported on the columns, whereas row labels show the proportion of nonzero entries in the generated matrix. The lighter the color, the larger the corresponding AUPRC value.}\label{fig:heatmap}
\normalsize
\end{center}
\end{figure}

Overall, these results suggests that MTLP should be preferred when the proportion of positives is drastically smaller than that of negatives. When data are more balanced, MTLP-inv better exploits the similarities among tasks and, at least in term of AUPRC, is a valid option. In terms of multilabel accuracy, MTLP is always better than MTLP-inv. Finally, it is worth noting that both methods outperforms LP in terms of AUPRC (see Section~\ref{sub:soa_comp} for LP results), whereas in terms of $\Fmax$ only MTLP achieves better results than LP.
%%%%%%%%%%%%%%%%%%%%%%%%%%%%%%%%%%%%%%%%%%%%%%%%%%%%%%%%%%%%
 \begin{table*}[t!]
     \begin{center}
         %%\vspace{-1cm}  
         %\begin{small}%\small   %Ered       %BIASred    %Net-VARred %UNBVARred  
         \small 
         {\centering 
             \begin{tabular}{@{}lcccc|cccc|cccc@{}} 
                 \hline
                 \hline\\[-5.5pt]
                 %\multicolumn{1}{|c|}{$\mathbf{Data set}$} & \multicolumn{5}{|c|}{$\mathbf{Methods}$} & \multicolumn{1}{|c|}{$\mathbf{Performance}$}\\
                 \texttt{METHODS} & \multicolumn{4}{c|}{\texttt{BP}} & \multicolumn{4}{c|}{\texttt{MF}}& \multicolumn{4}{c}{\texttt{CC}}\\[1pt]                
                 &{\scriptsize {\texttt{All}}} & {\scriptsize \texttt{$5$-$20$}} & {\scriptsize\texttt{$21$-$100$}} & \scriptsize\texttt{$\Fmax$} &\multicolumn{1}{c}{\scriptsize {\texttt{All}}} & {\scriptsize \texttt{$5$-$20$}} & {\scriptsize\texttt{$21$-$100$}}& \scriptsize\texttt{$\Fmax$}&\multicolumn{1}{c}{\scriptsize {\texttt{All}}} & {\scriptsize \texttt{$5$-$20$}} & {\scriptsize\texttt{$21$-$100$}}& \scriptsize\texttt{$\Fmax$}\\[1pt]
\hline\\[-5.5pt]
& \multicolumn{10}{c}{\texttt{FLY}}\\[2pt]
MTLP $\diss_0$    & \textbf{0.140}   & \textbf{0.133}  & \textbf{0.153}  & \textbf{0.247} & \textbf{\underline{0.333}}  & \textbf{0.322} & \textbf{0.355} & \textbf{0.411}  & \textbf{0.262} & \textbf{0.265} & \textbf{0.253} & 0.354 \\[2pt]                                       
 MTLP $\diss_3$    & \textbf{0.140} & \textbf{0.133} &  \textbf{0.153}  & {0.246} & \textbf{\underline{0.333}} & \textbf{0.322} & \textbf{0.355} & {0.410}  &  \textbf{0.262} & \textbf{0.265} & \textbf{0.253} & \textbf{0.357} \\[2pt]     
 MTLP-inv $\simm_1$    & 0.020 & 0.013 & 0.031 & 0.183 & 0.198  & 0.179 & 0.238 & 0.374 & 0.150  & 0.138 & 0.181 & 0.306 \\[2pt] 
 MTLP-inv $\simm_2$    & 0.020 & 0.014 & 0.031 & 0.170  & 0.192 & 0.172 &  0.235  & 0.351 & 0.101 & 0.082 & 0.147 & 0.259 \\[2pt]
 MTLP-inv $\simm_3$    & 0.135 & 0.129 & 0.146 & 0.244  & 0.328 & 0.318 &  0.352  & 0.381  & 0.261 & 0.265 & 0.251 & 0.333 \\[2pt] 
  %%%%%%%%%%%%%%%%%%%%%%%%%%%%%%%%%%%%%%%%%%%%%%%%%%%%%%%%%%%%%%%%%%%%%%%     
& \multicolumn{10}{c}{\texttt{HUMAN}}\\[2pt]
MTLP $\diss_0$    & {0.144}   & 0.133  & \textbf{0.165}  & 0.273 & {0.248}  & {0.247} & \textbf{0.250} & 0.383 & \textbf{0.224} & \textbf{0.259} & {0.156} & 0.317 \\[2pt]                                       
 MTLP $\diss_3$    & \textbf{\underline{0.145}} & \textbf{0.134}  & \textbf{0.165}  & \textbf{0.275} & \textbf{\underline{0.249}} & \textbf{0.248} & \textbf{0.250} & \textbf{0.385}  &  \textbf{0.224} & \textbf{0.259} & 0.156 & \textbf{0.318} \\[2pt]     
 MTLP-inv $\simm_1$    & 0.008 & 0.005 & 0.014 & 0.200 & 
 						0.093  & 0.083 & 0.152 & 0.330 & 
 						0.105  & 0.113 & 0.090 & 0.274 \\[2pt] 
 MTLP-inv $\simm_2$    & 0.008 & 0.005 & 0.012 & 0.182  
 					  & 0.059 & 0.050 &  0.079  & 0.294 
 					  & 0.066 & 0.064 & 0.068 & 0.223 \\[2pt]
 MTLP-inv $\simm_3$    & 0.139 & 0.129 & 0.159 & 0.244  & 0.243 & 0.241  &  0.244  & 0.355  & 0.220 & 0.256 & \textbf{0.160} & 0.299 \\[2pt]  
  %%%%%%%%%%%%%%%%%%%%%%%%%%%%%%%%%%%%%%%%%%%%%%%%%%%%%%%%%%%%%%%%%%%%%%%     
& \multicolumn{10}{c}{\texttt{BACTERIA}}\\[2pt]
MTLP $\diss_0$    & {0.119}   & {0.107}  & 0.169  & 0.210 & {0.173}  & {0.157} & 0.238 & 0.269 & {0.122} & {0.105} & {0.220} & \textbf{0.348} \\[2pt]                                       
 MTLP $\diss_3$    & 0.119 & {0.107} &  {0.168}  & \textbf{0.212} & {0.173} & {0.157} & {0.238} & \textbf{0.276}  &  {0.122} & 0.105 & \textbf{0.219} & \textbf{0.348} \\[2pt]     
 MTLP-inv $\simm_1$    & 0.069 & 0.056 & 0.123 & 0.181 & 0.106  & 0.092 & 0.165 & 0.235 & 0.101  & 0.086 & 0.187 & 0.246 \\[2pt] 
 MTLP-inv $\simm_2$    & 0.053 & 0.043 & 0.094 & 0.109  & 0.057 & 0.045 &  0.107  & 0.117 & 0.106 & 0.089 & 0.207 & 0.281 \\[2pt]
 MTLP-inv $\simm_3$    & \textbf{0.121} & \textbf{0.108} & \textbf{0.176} & 0.189  & \textbf{\underline{0.181}} & \textbf{0.165} &  \textbf{0.247}  & 0.247  & \textbf{0.123} & \textbf{0.107} & 0.212 & 0.289 \\[2pt]
 \hline
 \hline
 \end{tabular}
\normalsize}
            %\end{small}
\end{center}
\caption{Comparison according to average AUPRC and multilabel F-measure ($\Fmax$) between MTLP and MTLP-inv using the semantic similarity measures described in Section~\ref{subsub:simmat}. Column \texttt{All} is the average across all GO terms, column $5$-$20$ is the average across GO terms with at most $20$ positive genes, and column $21$-$100$ is the average across terms with more than $20$ positives. Best results are in boldface. Results are underlined when the difference between MTLP and MTLP-inv is statistically significant (Wilcoxon signed rank test, $p$-$value < 0.05$).}\label{tab:simmeas_comp}    
\end{table*}
In order to investigate the reasons why, unlike MTLP-inv, MTLP performance slightly varies with the task dissimilarity measure, we run MTLP on the \textit{fly} organism and CC tasks by randomly generating the matrix $\bar\bscC$. We generated matrices with different sparsity (from $5\%$ to $95\%$, with steps of $10\%$) and with different ranges of weight values. Specifically, we uniformly selected weights in the interval $[0,\tau]$, with $\tau$ ranging from $0.1$ to $1$, by steps of $0.1$. In Figure~\ref{fig:heatmap}, we show the heatmap of the average AUPRC obtained in each experiment. 
%%%%%%%%%%%%%%%%%%%%%%%%%%%%%%%%%%%%%%%%%%%%%%%%%%
As expected, the results are considerably worse than those obtained when considering real dissimilarity matrices (see Table~\ref{tab:simmeas_comp}). There is a small AUPRC variation from the different random data, with higher AUPRC when the dissimilarity matrix is denser and with larger entries (the former seems to affect the results more than the latter). This is consistent with Fact~\ref{th:1}, since the lower the weight and/or the sparser the matrix, the closer MTLP is to LP.
Finally, on randomly generated dissimilarity matrices MTLP performs even worse than LP, as we can see from Figure~\ref{fig:fly_homo__ecoli_comp}. 

%This also partially explains why the performance of MTLP on \textit{m1}/\textit{m2} and \textit{m3} matrices are indistinguishable: on the one hand, the percentage of non-zero entries in matrices \textit{m1}/\textit{m2} and \textit{m3} is the almost the same, $0.99725$ and $0.99724$ respectively; on the other hand, the average weights are close: $0.997$ (standard deviation $0.029$) and $0.827$ (standard deviation $0.112$) respectively for \textit{m1}/\textit{m2} and \textit{m3}. Another explanation for the limited variability of the results of MTLP on the different dissimilarity matrices adopted is given by Fact~\ref{th:1}. Indeed, the effect of transformation $\psi_{\bar\bscA}$ on a given instance $i$ for task $k$ depends on the positive $\mathfrak{d}^+_{k,i}$ and negative $\mathfrak{d}^-_{k,i}$ multi-task degree of $i$, and when both sparsity and weight range in the matrix $\bar\bscC$ are close, even the positive and negative degree are likely to vary just slightly.   
%%%%%%%%%%%%%%%%%%%%%%%%%%%%%%%%%%%%%%%%%%%%%%%%%%%%%%%%%%%%   
\begin{figure}[!ht]
\begin{center}
\scriptsize
\begin{tabular}{c}
\hspace{-0.5cm} \includegraphics[angle=0, width=0.41\textwidth]{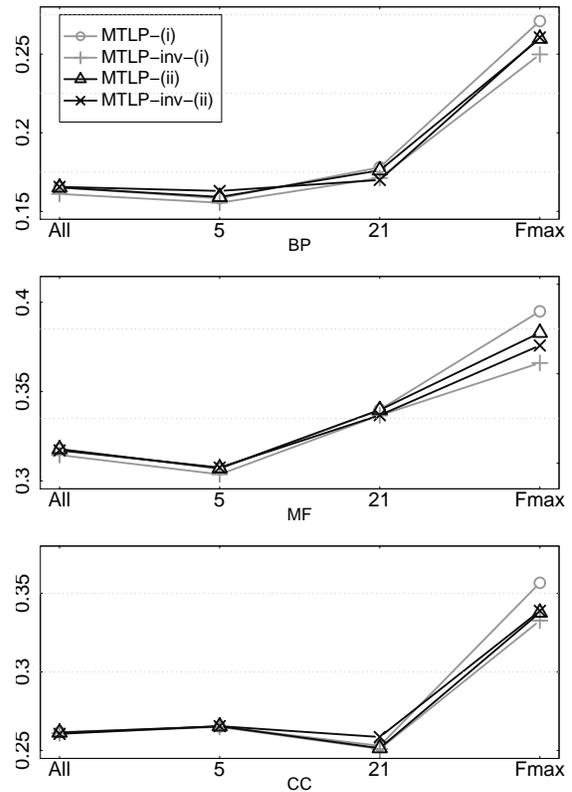}	 
\end{tabular}
\caption{Average AUPRC performance across all GO terms (All), across GO terms with at most $20$ positive instances ($5$), and across terms with more than $20$ positives ($21$). }\label{fig:grouping_strategies}
\normalsize
\end{center}
\end{figure}
%%%%%%%%%%%%%%%%%%%%%%%%%%%%%%%%%%%%%%%%%%%%%%%%%%%%%%%%
%%%%%%%%%%%%%%%%%%%%%%%%%%%%%%%%%%%%%%%%%%%%%%%%%%%%%%%%%%%%%%%
\subsubsection{Grouping GO terms for multitask mapping}\label{sub:go_group}
%%%%%%%%%%%%%%%%%%%%%%%%%%%%%%%%%%%%%%%%%%%%%%%%%%%%%%%%%%%%
Following the approach proposed in~\cite{Mostafavi10}, in addition to the strategy grouping GO terms by branch (i) adopted in the previous section, we have examined an alternative way for grouping the terms to be considered in the multitask maps (\ref{eq:invmap}) and (\ref{eq:map_over_A}) when running MTLP-inv and MTLP algorithms, respectively.
Specifically, we grouped GO terms not just by GO branch (BP, MF, and CC), but also by taking into account the number of annotated proteins (ii), obtaining $6$ groups: BP with $5$-$20$ ($1119$ terms) and $21$-$100$ ($623$ terms) annotations, MF $5$-$20$ ($362$ terms), $21$-$100$ ($177$ terms), and CC $5$-$20$ ($267$ terms), $21$-$100$ ($109$ terms).   
The corresponding results on \textit{fly} data are reported in Figure~\ref{fig:grouping_strategies}.
AUPRC results show negligible differences between strategies (i) and (ii), for both MTLP and MTLP-inv. More clear is the difference in terms of \textit{Fmax}, with opposite behaviour between MTLP and MTLP-inv: MTLP has worse performance in all GO branches; MTLP-inv instead tends to perform better (see for instance MF results). Indeed, black lines (grouping strategy (ii)) in correspondence of \textit{Fmax} are always between grey lines (grouping strategy (i)). However, the best results are still achieved by MTLP when grouping terms by GO branch, and accordingly we consider this strategy in the rest of the paper. 
%%%%%%%%%%%%%%%%%%%%%%%%%%%%%%%%%%%%%%%%%%%%%%%%%%%%%%%%%%%%%%%%%%%%%%%%%%%%%%%%%%%% 
\subsubsection{Prediction of GO functions for fly, human, and bacteria organisms}\label{sub:soa_comp}
MTLP ($\bar\gamma = 1$) was compared with state-of-the-art graph-based methodologies applied to the prediction of protein functions. We considered: \textit{LP}, the label propagation algorithm described in Section~\ref{subsec:LP}; \textit{COSNetM}~\cite{Frasca15}, an extension of a node classifier designed for unbalanced settings~\cite{Bertoni11}; \textit{RW}, the classical $t$-step random walk algorithm~\cite{Lovasz96}; \textit{GBA}, a method based on the \textit{guilt-by-association} assumption~\cite{Schwikowski00}; \emph{MS-kNN}, one of the best methods in the recent CAFA2 challenge applying the \textit{kNN} algorithm to each network independently, and then combining the obtained predictions~\cite{Lan13}. 

In order to deal with label imbalance in LP, we applied a label normalization step before running label propagation. This step normalizes the labels of each GO term so that positive and negative labels sum to $1$. In our experiments, this variant of LP performs much better than the vanilla LP algorithm.
For the RW algorithm we set the limit on the number of iterations to $100$, since higher values did not improve the performance while increasing the computational burden. Finally, we set to $5$ the parameter $k$ for the kNN algorithm, as a result of a tuning process on training data.
%%%%%%%%%%%%%%%%%%%%%%%%%%%%%%%%%%%%%%%%%%%%%%%%%%%%%%%%%%%%   
\begin{figure*}[!th]
\begin{center}
\scriptsize
\begin{tabular}{c}
\hspace{-0cm} \includegraphics[angle=-90, width=0.9\textwidth]{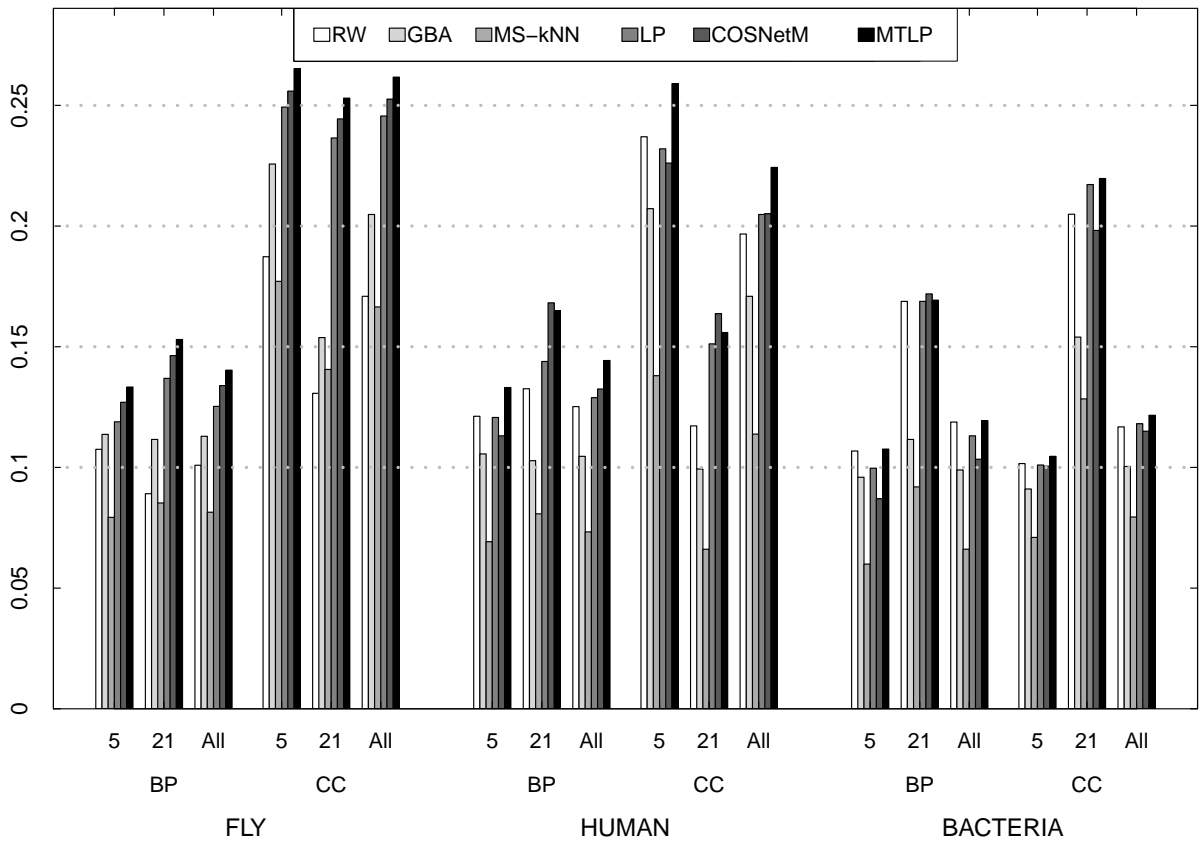}	 
\end{tabular}
\caption{Average AUPRC performance across all GO terms (All), across GO terms with at most $20$ positive instances ($5$), and across terms with more than $20$ positives ($21$). }\label{fig:fly_homo__ecoli_comp}
\normalsize
\end{center}
\end{figure*}
%%%%%%%%%%%%%%%%%%%%%%%%%%%%%%%%%%%%%%%%%%%%%%%%%%%%%%%%
%%%%%%%%%%%%%%%%%%%%%%%%%%%%%%%%%%%%%%%%%%%%%%%%%%%%%%%%%%%%   
\begin{figure*}[!th]
\begin{center}
\scriptsize
\begin{tabular}{c}
\hspace{-0cm} \includegraphics[angle=-90,width=0.8\textwidth]{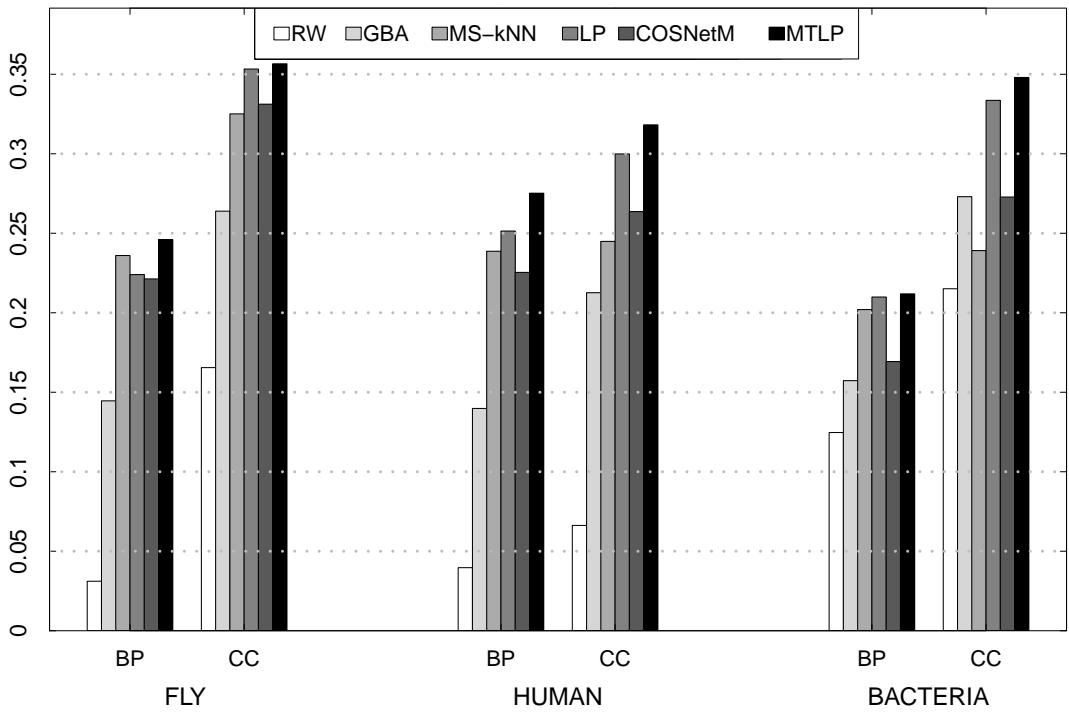}	 
\end{tabular}
\caption{Average multi-label F-measure performance across all GO terms.}\label{fig:fly_homo__ecoli_F}
\normalsize
\end{center}
\end{figure*}
%%%%%%%%%%%%%%%%%%%%%%%%%%%%%%%%%%%%%%%%%%%%%%%%%%%%%%%%

In Figures~\ref{fig:fly_homo__ecoli_comp} and~\ref{fig:fly_homo__ecoli_F} we show the obtained results in terms of AUPRC and $\Fmax$, on BP and CC terms respectively (on MF terms the methods showed a similar behaviour).
Interestingly, MTLP always achieves the highest AUPRC averaged over all tasks (\textit{All}), with statistically significant improvements over the second top method ($p$-$value < 0.001$), except for \textit{bacteria} data and for BP terms on \textit{fly} data. When comparing with LP method, the improvement is always significant, except for CC (\textit{bacteria} data). COSNetM is the second method on \textit{human} and \textit{fly} data sets, while on \textit{bacteria} LP (CC) and RW (BP) rank as second method.
Furthermore, and more importantly, MTLP improvements are more noticeable on the most unbalanced terms, which are those best characterizing the biological functions of genes.
GBA, MS-kNN and RW methods seem suffer the strongly unbalanced setting, and perform worse than LP, with the exception of RW on \textit{bacteria} data set. The good performance of COSNetM in this unbalanced setting is likely due to its cost-sensitive strategy, which requires learning two model parameters. This extra learning step increases its computation time. Indeed, COSNetM takes on average around $4$ seconds on a Linux machine with Intel Xeon(R) CPU 3.60GHz and 32 Gb RAM to perform an entire cross validation cycle for one task on \textit{fly} data, whereas both LP and MTLP take on average slightly less than one second. This confirms our observation that applying the map $\psi_{\bar\bscA}$ after label propagation does not increase the algorithm complexity, and just slightly increases the execution time for computing $\psi_{\bar\bscA}$.

Even in terms of \textit{Fmax} MTLP obtains the best results, with LP second-best method (except on BP --- \textit{fly} data). This shows that our method can achieve good predictive capabilities both when predicting single GO terms and when predicting a GO multilabel for single proteins. On the other side, the compared methods tend to have competitive performance in only one scenario. For instance, RW poorly performs in terms of $\Fmax$, whereas, unlike AUPRC, MS-kNN achieves good \textit{Fmax} results: on BP (\textit{fly} data) it is the best method after MTLP. Even COSNetM, which is the second method in terms of AUPRC, achieves the third or the fourth best $\Fmax$ rank.
%%%%%%%%%%%%%%%%%%%%%%%%%%%%%%%%%%%%%%%%%%%%%%%%%%%%%%%%%%%%%
\subsubsection{Evaluating different powers of the Laplacian matrix}\label{subsub:p}
\begin{figure}[!h]
\begin{center}
\scriptsize
\begin{tabular}{c}
\hspace{-0.5cm} \includegraphics[angle=-90, width=0.45\textwidth]{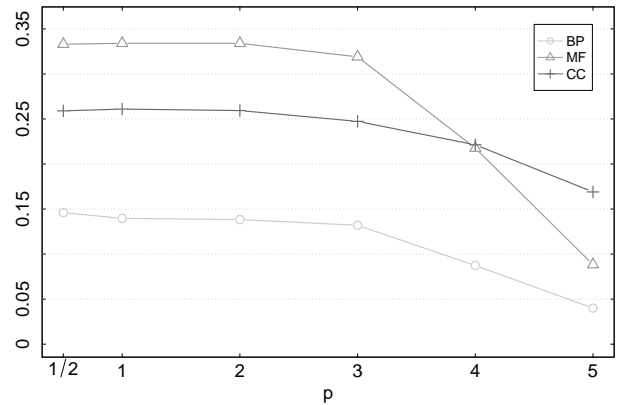}	 
\end{tabular}
\caption{Average AUPRC values achieved by MTLP of \textit{fly} data with different values of the parameter $p$.}\label{fig:pTuning}
\normalsize
\end{center}
\end{figure}

A further experiment was carried out to analyze how MTLP performance changes when using the map $\psi_{\bar\bscA,p}(\bY) = \bY \bar\bscA^p$ for $p \ge \tfrac{1}{2}$, instead of $\psi_{\bar\bscA}(\bY) = \bY \bar\bscA$. We empirically tested on the \textit{fly} organism different values of $p$, fixing the parameter $\bar\gamma = 1$ and using the $\diss_3$ measure. The results are shown in Figure~\ref{fig:pTuning}. We considered $p=\tfrac{1}{2}, 2, 3, 4, 5$. Except for BP terms, where the map $\psi_{\bar\bscA,1/2}$ performs slightly better than $\psi_{\bar\bscA,1}$, all choices of $p \neq 1$ lead to worse results. In particular, the performance strongly decays for $p > 2$.

%%%%%%%%%%%%%%%%%%%%%%%%%%%%%%%%%%%%%%%%%%%%%%%%%%%%%%%%%%%%%%%%
\subsubsection{Impact of parameter $\bar\gamma$}\label{subsub:gamma}
Large values of the $\bar\gamma$ parameter, introduced in Section~\ref{sec:MTLP}, tend to reduce the multitask contribution encoded in $\bar\bscA$, since $\bar\bscA$ is diagonally dominant and absolute labels assigned to positives and negatives vertices by the map $\psi_{\bar\bscA}$ tend to be almost the same (see Fact~\ref{th:1}).
Hence, this allows to ``regulate'' to some extent the method between multitask and singletask label propagation.
We experimentally tuned $\bar\gamma$ on \textit{fly} and \textit{human} data from $0.25$ to $1.5$ with step size $0.25$. It turns out there is a negligible difference, with results reported in Table~\ref{tab:simmeas_comp} and corresponding to $\bar\gamma = 1$. This is expected, since $m$ is much larger than $1$ in the considered experiments. For this reason, we also performed another experiment in which we selected a smaller subset of terms in the BP branch (a similar trend is observed for the MF and CC branches). Specifically, we ran our algorithm on a subset of $42$ terms for the \textit{fly} organism, by varying $\bar\gamma$ in the specified range. The results are shown in Table~\ref{tab:gammaTuning}. Confirming our observations, MTLP is more sensitive to $\bar\gamma$ values in this setting, and the overall trend is that the average AUPRC tends to decrease when $\bar\gamma$ becomes larger (similarly to $\Fmax$). This not surprising: as we explained, with large values of $\bar\gamma$ MTLP behaves closer to LP, whose results are lower in this setting.

%%%%%%%%%%%%%%%%%%%%%%%%%%%%%%%%%%%%%%%%%%%%%%%%%%%%%%%%%%%%
\begin{table}[!t]
\begin{center}
\small
%%\vspace{-1cm}  
%\begin{small}%\small   %Ered       %BIASred    %Net-VARred %UNBVARred   
{\centering 
\begin{tabular}{@{}lccc@{}} 
\hline
\hline\\[-5.5pt]
{\texttt{$\bar\gamma$}} & {\texttt{All}} & {\texttt{$5$-$20$}}& {\texttt{$20$-$100$}}\\[1pt]                 
\hline\\[-5.5pt]
0.25      &  0.158  & 0.150 & 0.182  \\[2pt]     
0.5       & 0.157   & 0.151 & 0.177  \\[2pt]     
0.75     & 0.145   & 0.134 & 0.178  \\[2pt]     
1          &  0.144  & 0.133 & 0.178  \\[2pt]     
1.25     &  0.140  & 0.130 & 0.175  \\[2pt]     
1.5       & 0.139   & 0.129 & 0.174  \\[2pt]     
\hline
\hline
\end{tabular}
\normalsize}
%\end{small}
\end{center}
\caption{AUPRC of the MTLP method ($p=1$, task similarity measure $\diss_3$ averaged across $42$ selected MF GO terms for \textit{human} data by varying the parameter $\bar\gamma$. Column \texttt{All} is the average across all tasks, column $5$-$20$ is the average across terms with at most $20$ annotations, and column $21$-$100$ is the average across terms with more than $20$ positives.}
\label{tab:gammaTuning}    
\end{table}

\section{Conclusions}
We have shown that task relatedness information represented through task dissimilarity is better suited for label propagation in unbalanced protein function prediction than task similarity. The proposed multitask label propagation algorithm compared favourably with the state-of-the-art methodologies for protein function prediction on three model organisms.
Although we gained some intuition and collected empirical evidence, we are still investigating the multitask problems where our approach is most effective. Specifically, it would be useful to study whether dissimilarity information helps when coupled with multitask algorithms different from label propagation. For example, linear learning algorithms such as SVM or Perceptron.
Laplacian spectral theory is also likely to help us shed some further light on the properties of our method.

% The very first letter is a 2 line initial drop letter followed
% by the rest of the first word in caps (small caps for compsoc).
% 
% form to use if the first word consists of a single letter:
% \IEEEPARstart{A}{demo} file is ....
% 
% form to use if you need the single drop letter followed by
% normal text (unknown if ever used by the IEEE):
% \IEEEPARstart{A}{}demo file is ....
% 
% Some journals put the first two words in caps:
% \IEEEPARstart{T}{his demo} file is ....
% 
% Here we have the typical use of a "T" for an initial drop letter
% and "HIS" in caps to complete the first word.
%\IEEEPARstart{T}{his} demo file is intended to serve as a ``starter file''
%for IEEE Computer Society journal papers produced under \LaTeX\ using
%IEEEtran.cls version 1.8b and later.
% You must have at least 2 lines in the paragraph with the drop letter
% (should never be an issue)
%I wish you the best of success.

%\hfill mds
 
%\hfill August 26, 2015

% use section* for acknowledgment
\ifCLASSOPTIONcompsoc
  % The Computer Society usually uses the plural form
  \section*{Acknowledgments}
\else
  % regular IEEE prefers the singular form
  \section*{Acknowledgment}
\fi

The authors would like to thank the reviewers of BIOKDD16 for useful comments on an earlier draft of this paper.

% Can use something like this to put references on a page
% by themselves when using endfloat and the captionsoff option.
\ifCLASSOPTIONcaptionsoff
  \newpage
\fi

\begin{IEEEbiography}[{\includegraphics[width=1.1in,height=1.25in,clip,keepaspectratio]{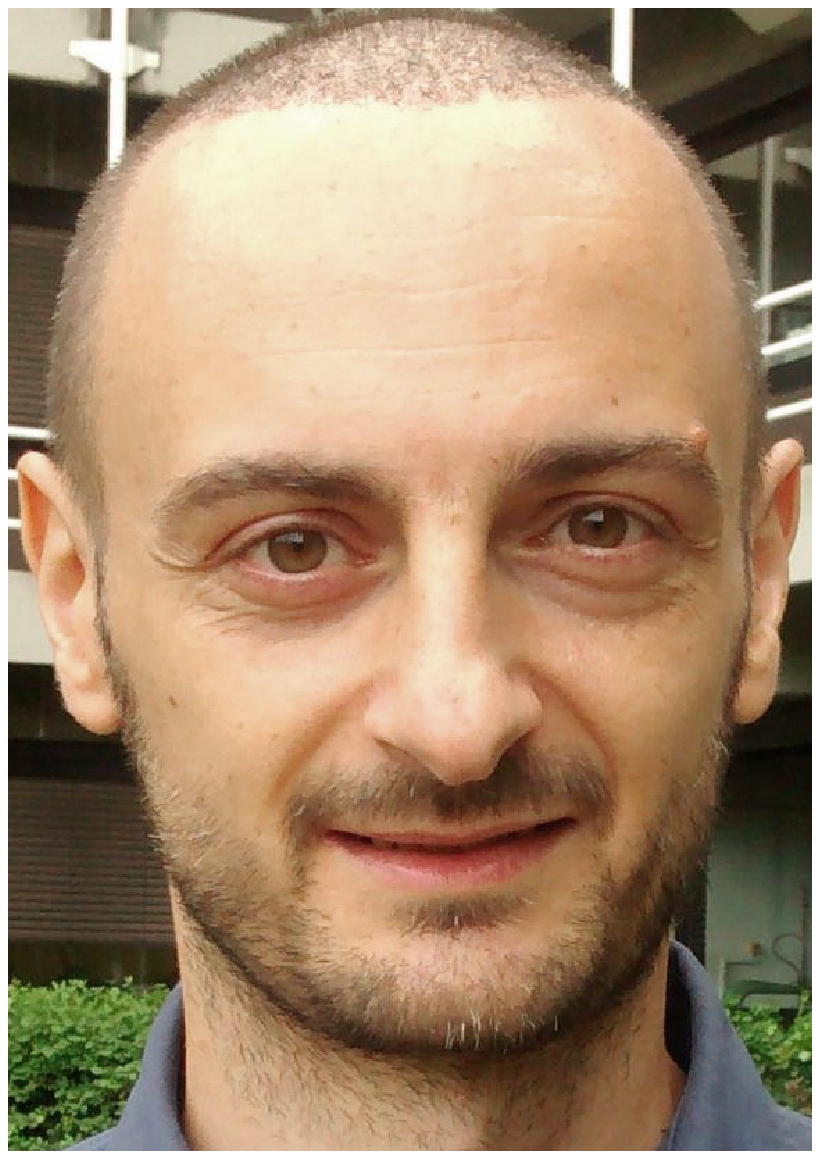}}]{Marco Frasca}
received the Ph.D. degree in Computer Science from University of Milano, Italy in 2012. He is currently a post-doc research fellow in Computer Science at the University of Milano. His research interests include the study of neural networks models for unbalanced classification problem and the development of machine learning techniques for emerging problems life sciences, such as protein function prediction, gene-disease prioritization and drug repositioning. 
\end{IEEEbiography}

% if you will not have a photo at all:
\begin{IEEEbiography}[{\includegraphics[width=1.1in,height=1.25in,clip,keepaspectratio]{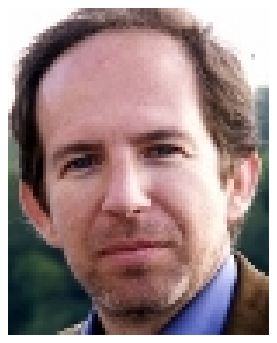}}]{Nicol\`{o} Cesa Bianchi}
is professor of Computer Science at the University of Milano, Italy. He held visiting positions with UC Santa Cruz, Graz Technical University, Ecole Normale Superieure (Paris), Google, and Microsoft Research. He received a Google Research Award and a Xerox University Affairs Committee Award. His research interests include theory and applications of machine learning, sequential optimization, and algorithmic game theory. On these topics, he published two monographs: \textsl{Prediction, Learning, and Games} and \textsl{Regret Analysis of Stochastic and Nonstochastic Multi-armed Bandit Problems}.
\end{IEEEbiography}

% insert where needed to balance the two columns on the last page with
% biographies
%\newpage

% You can push biographies down or up by placing
% a \vfill before or after them. The appropriate
% use of \vfill depends on what kind of text is
% on the last page and whether or not the columns
% are being equalized.

%\vfill

% Can be used to pull up biographies so that the bottom of the last one
% is flush with the other column.
%\enlargethispage{-5in}

% that's all folks
\end{document}